\documentclass[12pt,fleqn]{article}

\usepackage{amsmath, amssymb}
\usepackage{natbib}
\usepackage{fancyhdr, graphicx}
\usepackage{color}
\usepackage{float} 	%% Use to fix Figure or Table: ex: \begin{table}[H]

\title{BURGERS' PINNs WITH IMPLICIT EULER TRANSFER LEARNING}

\author
    {\rm \begin{tabular}{l}
    \textbf{Vitória Biesek} - \texttt{vitoriabiesek@gmail.com}\\%
    \textbf{Pedro H. A. Konzen} - \texttt{pedro.konzen@ufrgs.br}\\
    {\fontsize{11}{0}\selectfont UFRGS-IME, Porto Alegre, Brazil}
  \end{tabular}}

\usepackage{booktabs}
\newcommand{\e}{\mathrm{E}}

\begin{document}

\maketitle

\begin{abstract}
  The Burgers equation is a well-established test case in the computational modeling of several phenomena such as fluid dynamics, gas dynamics, shock theory, cosmology, and others. In this work, we present the application of Physics-Informed Neural Networks (PINNs) with an implicit Euler transfer learning approach to solve the Burgers equation. The proposed approach consists in seeking a time-discrete solution by a sequence of Artificial Neural Networks (ANNs). At each time step, the previous ANN transfers its knowledge to the next network model, which learns the current time solution by minimizing a loss function based on the implicit Euler approximation of the Burgers equation. The approach is tested for two benchmark problems: the first with an exact solution and the other with an alternative analytical solution. In comparison to the usual PINN models, the proposed approach has the advantage of requiring smaller neural network architectures with similar accurate results and potentially decreasing computational costs.

\end{abstract}

\section{INTRODUCTION}
Burgers equation appears in the modeling of several phenomena such as fluid dynamics, gas dynamics, shock theory, cosmology, and others \citep{Bonkile2018a}. It is a well-established test case in mathematical analysis and numerical simulations of convective-diffusive partial differential equations \citep{Konzen2017a}. We consider the viscous Burgers equation with initial and homogeneous Dirichlet boundary conditions
\begin{subequations}\label{eq:burgers-prob}
  \begin{align}
    &u_t + uu_x = \nu u_{xx}, (t, x)\in (0, t_f]\times (0, 1) \label{eq:burgers}\\
    &u(0, x) = u_0(x), x\in [0, 1], \label{eq:burgers-ic}\\
    &u(t, 0) = u(t, 1) = 0, t\in [0, t_f], \label{eq:burgers-bc}
\end{align}
\end{subequations}
with given $u_0 = u_0(x)$. For instance, in a fluid dynamics application, it models the velocity $u = u(t,x)$ $(\text{m}/\text{s})$ at time $t$ $(\text{s})$ and point $x$ $(\text{m})$, of a fluid with kinematic viscosity $\nu$ $(\text{m}^2/\text{s})$.

In this work, we investigate the application of Physics-Informed Neural Networks (PINNs) with an implicit Euler transfer learning approach to solve the Burgers equation. PINNs \citep{Raissi2019a} are Deep Learning \citep{Goodfellow2016a} techniques to solve Partial Differential Equations (PDEs). Recently, they have been applied to solve many important problems, such as incompressible Navier-Stokes equations \citep{Raissi2018a, Jin2021a}, Euler equations for high-speed aerodynamic flows \citep{Mao2020a}, heat transfer problems \citep{Cai2021a}, and the advection equation \citep{Vadyala2022a}.

For a time-dependent PDE, the PINN consists of a Multilayer Perceptron (MLP) (\citep{Haykin2008a}) neural network that learns the solution $u(t,x)$ from the governing equations. Usually, the MLP has inputs $(t,x)$ and outputs $u(t,x)$, or has input $x$ and outputs $\pmb{u}(x) = \left\{u\left(t^{(k)},x\right)\right\}_{k=0}^{n_t}$ for given $n_t$ time steps $t^{(k)}$. Here, we investigate the application of an alternative PINN approach, in which the solution is given by a sequence of MLPs, one per time step $t^{(k)}$. From the initial condition, a first MLP $\mathcal{N}^{(0)}$ learns the solution at $t^{(0)} = 0$ by training to approximate the function $u_0 = u_0(x)$. Then, given a time step size $h_t>0$, the model $\mathcal{N}^{(0)}$ transfers its knowledge to a second MLP $\mathcal{N}^{(1)}$, which learns the solution at $t^{(1)} = h_t$ by training on the differential problem that arises from the implicit Euler scheme of the Burgers equation. It is an iterative procedure, the $k$-th $\mathcal{N}^{(k)}$ model transfers its knowledge to initiate the $\mathcal{N}^{(k+1)}$ model, which learns the solution at time $t^{(k+1)}$ from an implicit Euler scheme.

The proposed approach allows us to deal with smaller neural network architectures in comparison to the usual PINN approaches. This has the potential to enhance computational performance. In the following, the methodology is presented, two numerical test cases are discussed, and conclusions are then given.

\section{BURGERS' PINNs}
Physics-Informed Neural Networks (PINNs) are Deep Learning techniques to solve Partial Differential Equations (PDEs). The solution is sought by supervised learning tasks that embed the PDE problem into the loss function. Here, we describe an alternative PINN approach, where a sequence of MLPs, one for each discrete time step, is trained with an Euler implicit transfer learning scheme.

\subsection{Multilayer perceptron}
\begin{figure}[h!]
  \centering
  \includegraphics[width=0.8\textwidth]{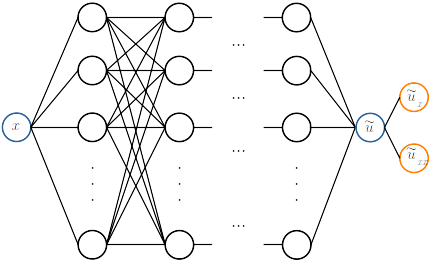}
  \caption{Multilayer Perceptron architecture $1 - n_n\times n_l - 1$ with automatic differentiation.}
  \label{fig:pinn}
\end{figure}

In this work, we considered Multilayer Percetrons (MLPs) of the form
\begin{equation}
  \tilde{u} = \mathcal{N}\left(x; \left\{\left(W^{(l)},\pmb{b}^{(l)},\pmb{f}^{(l)}\right)\right\}_{l=1}^{n_l}\right)
\end{equation}
where the triple $\left(W^{(l)},\pmb{b}^{(l)},\pmb{f}^{(l)}\right)$ denotes the weights $W^{(l)}$, the biases $\pmb{b}^{(l)}$ and the activation function $\pmb{f}^{(l)}$ in the $l$-th layer of the network, $l= 1, 2, \dotsc, n_l$, with $n_l$ a given number of layers (see Fig.~\ref{fig:pinn}). As a Deep Learning technique, the forward processing is computed by iterative compositions
\begin{equation}
  \pmb{a}^{(l)} = \pmb{f}^{(l)}\left(W^{(l)}\pmb{a}^{(l-1)}+\pmb{b}^{(l)}\right),
\end{equation}
with the input $a^{(0)} = x$, the output $a^{(n_l)} = \tilde{u}$ and $l = 1, 2, \dotsc, n_l$. Assuming the needed smoothness of $\pmb{f}^{(l)}$, the derivatives $\tilde{u}_x$ and $\tilde{u}_{xx}$ can be computed by automatic differentiation as an application of the chain rule.

\subsection{Implicit Euler transfer learning}
The implicit Euler time scheme to Eq.~\eqref{eq:burgers-prob} consists in the iteration
\begin{subequations}
  \begin{align}
    &\tilde{u}^{(0)}(x) = u_0(x),\label{eq:euler_ic}\\
    &\tilde{u}^{(k)} = \tilde{u}^{(k-1)} + h_t\left(\nu \tilde{u}^{(k)}_{xx} - \tilde{u}^{(k)}\tilde{u}^{(k)}_x\right),\\
    &\tilde{u}^{(k)}(0) = \tilde{u}^{(k)}(1) = 0,
  \end{align}
\end{subequations}
where $\tilde{u}^{(k)}(x) \approx u\left(t^{(k)}, x\right)$ at each time $t^{(k)} = kh_t$, $k = 0, 1, 2, \dotsc, n_t$, time step size $h_t=1/n_t$, for a given number of time steps $n_t$. The proposed PINN with implicit Euler transfer learning consists in training a sequence of MLPs
\begin{equation}
  \tilde{u}^{(k)}(x) = \mathcal{N}^{(k)}(x),
\end{equation}
where $k = 0, 1, 2, \dotsc, n_t$, with the input $x$ and output the estimation of $\tilde{u}^{(k)}$.

From the initial condition Eq.~\eqref{eq:euler_ic}, the neural network $\mathcal{N}^{(0)}$ is trained to minimize the loss function
\begin{equation}\label{eq:loss_ic}
  \mathcal{L}_0 := \frac{1}{n_s}\sum_{s=1}^{n_s}\left|\tilde{u}^{(0)}(x_s) - u_0(x_s)\right|^2,
\end{equation}
with $n_s$ samples $0 \leq x_s \leq 1$. Sequentially, $k = 1, 2, \dotsc, n_t$, the knowledge of the neural network $\mathcal{N}^{(k-1)}$ is transferred to $\mathcal{N}^{(k)}$, which is then trained to minimize the loss function
\begin{equation}\label{eq:loss}
  \mathcal{L} := \frac{1}{n_s-2}\sum_{s=1}^{n_s-2}\left|\mathcal{R}\left(x; \tilde{u}^{(k)},\tilde{u}^{(k-1)}\right)\right|^2 + \frac{1}{2}\left(\left|\tilde{u}^{(k)}(0)\right|^2 + \left|\tilde{u}^{(k)}(1)\right|^2\right),
\end{equation}
where $\mathcal{R}$ denotes the residual
\begin{equation}
  \mathcal{R}\left(x; \tilde{u}^{(k)},\tilde{u}^{(k-1)}\right) := \tilde{u}^{(k)} - \tilde{u}^{(k-1)} - h_t\left(\nu \tilde{u}^{(k)}_{xx} - \tilde{u}^{(k)}\tilde{u}^{(k)}_x\right).
\end{equation}
The derivatives are directly computed from the neural network model by automatic differentiation.

The basic algorithm of the proposed PINN with implicit Euler transfer learning is summarized as follows:
\begin{itemize}
\item[0.] Set the number $n_t$ and the size $h_t$ of time steps.
\item[1.] Set $\mathcal{N}^{(0)}$ architecture.
\item[2.] Train $\mathcal{N}^{(0)}$ to minimize the initial loss function Eq.~\eqref{eq:loss_ic}.
\item[3.] For $k = 1, 2, \dotsc, n_t$:
  \begin{itemize}
  \item[a.] (Knowledge transfer.) $\mathcal{N}^{(k)} \leftarrow \mathcal{N}^{(k-1)}$.
  \item[b.] Train $\mathcal{N}^{(k)}$ to minimize the loss function Eq.~\eqref{eq:loss}.
  \end{itemize}
\end{itemize}
At the end, the approach provides the sequence of MLPs $\left\{\mathcal{N}^{(k)}\right\}_{k=0}^{n_t}$, each giving the estimated solution $\tilde{u}^{(k)}(x) \approx u\left(t^{(k)},x\right)$ of the Burgers equation \eqref{eq:burgers-prob}. Usually, there is no need to store the whole sequence, and the algorithm needs to store just $\mathcal{N}^{(k)}$ and $\mathcal{N}^{(k-1)}$ at each iteration.

\subsection{Implementation details}
We have performed Python implementations of the neural network models with the help of the PyTorch library \citep{Stevens2020a}. All the models considered are MLPs with an architecture $1 - n_n\times n_l - 1$ (see Fig.~\ref{fig:pinn}), one input, $n_l$ hidden layers each with $n_n$ units, and one output. In the hidden layers, the hyperbolic tangent is used as the activation function, and in the output layer, the identity function. The models are trained following a mini-batch gradient descent approach with the Adam method as the optimizer. As a training stop criteria, we have assumed $\mathcal{L}_0, \mathcal{L} < 10^{-6}$ (computation has been performed using float $32$-bits arithmetic).

\section{RESULTS}
Here, the proposed PINN with an implicit Euler knowledge transfer is tested for two benchmark problems.

\subsection{Problem 1}
The first problem has initial condition
\begin{equation}\label{eq:prob1_ic}
  u_0(x) = 2\nu\pi\frac{\sin(\pi x)}{2 + \cos(\pi x)}
\end{equation}
and exact solution \citep{Wood2006a}
\begin{equation}
  u(t,x) = 2\nu\pi\frac{e^{-\nu\pi^2 t}\sin(\pi x)}{2 + e^{-\nu\pi^2 t}\cos(\pi x)}.
\end{equation}
In the following, $\nu = 1$ is assumed.

We consider MLPs with structure $1-n_n\times n_l-1$, where $n_n$ is the number of units per layer $l = 1, 2, \dotsc, n_l$. In order to choose the $n_n$ and $n_l$, numerical tests have been performed for several choices of these parameters. Due to the stochasticity of the procedure (initialization and training algorithm), each test has been repeated three times. Table \ref{tab:prob1_ic} shows $\bar{n}_e/\bar{\mathcal{L}}_0$, the average number of epochs $\bar{n}_e$ (training iterations), and the final loss function value $\bar{\mathcal{L}}_0$. From these tests, we have concluded that a MLP $1-30\times 3 - 1$ with $n_s = 100$ was enough to learn the initial condition Eq.~\eqref{eq:prob1_ic}. With less than $10000$ epochs, the training is successfully ended to the chosen tolerance $\mathcal{L}_0 < 10^{-6}$.

\begin{table}[h!]
  \centering
  \caption{Problem 1. Model training tests of the initial condition Eq.~\eqref{eq:prob1_ic}. The entries $\bar{n}_e/\bar{\mathcal{L}}_0$ are the average number of epochs $\bar{n}_e$ (training iterations), and the final loss function value $\bar{\mathcal{L}}_0$.}
  \vspace{12pt}
  \begin{tabular}{c|cccc}\toprule
    $n_l$\textbackslash $n_n$ & 10             & 20                  & 30                  & 40 \\\midrule
    \multicolumn{5}{c}{$ns = 10$}\\\midrule
    $1$ & $5\e{+4}~/~2\e{-5}$ & $5\e{+4}~/~3\e{-5}$ & $5\e{+4}~/~6\e{-6}$ & $4\e{+4}~/~3\e{-5}$ \\
    $2$ & $1\e{+4}~/~6\e{-7}$ & $1\e{+4}~/~7\e{-7}$ & $1\e{+4}~/~7\e{-7}$ & $1\e{+4}~/~5\e{-7}$ \\
    $3$ & $1\e{+4}~/~6\e{-7}$ & $1\e{+4}~/~6\e{-7}$ & $9\e{+3}~/~8\e{-7}$ & $9\e{+3}~/~5\e{-7}$ \\
    $4$ & $1\e{+4}~/~8\e{-7}$ & $1\e{+4}~/~6\e{-7}$ & $2\e{+4}~/~8\e{-7}$ & $1\e{+4}~/~7\e{-7}$ \\\midrule
    \multicolumn{5}{c}{$ns = 100$}\\\midrule
    $1$ & $5\e{+4}~/~6\e{-6}$ & $5\e{+4}~/~5\e{-6}$ & $5\e{+4}~/~7\e{-6}$ & $5\e{+4}~/~1\e{-5}$\\
    $2$ & $5\e{+4}~/~9\e{-7}$ & $1\e{+4}~/~9\e{-7}$ & $1\e{+4}~/~9\e{-7}$ & $1\e{+4}~/~9\e{-7}$\\
    $3$ & $2\e{+4}~/~9\e{-7}$ & $1\e{+4}~/~9\e{-7}$ & $1\e{+4}~/~9\e{-7}$ & $9\e{+3}~/~9\e{-7}$\\
    $4$ & $2\e{+4}~/~9\e{-7}$ & $1\e{+4}~/~9\e{-7}$ & $1\e{+4}~/~9\e{-7}$ & $1\e{+4}~/~9\e{-7}$\\\midrule    
  \end{tabular}
  \label{tab:prob1_ic}
\end{table}

The model $\mathcal{N}^{(k)}$ inherits its initial parameters from $\mathcal{N}^{(k-1)}$ and its training performance depends on the time step $h_t$ and the number of mesh samples $n_s$. In order to study the influence of these parameters, we have performed numerical tests for varying them. Each test has been repeated three times. Table~\ref{tab:prob1_fs} shows $\bar{n}_e/\bar{\varepsilon}_{\text{rel}}$, the average of the number of epochs $\bar{n_e}$ and the final value of the $L^2$ relative error
\begin{equation}
  \varepsilon_{\text{rel}}\left(t^{(k)}\right) := \frac{\left\|\tilde{u}^{(k)}(x) - u\left(t^{(k)},x\right)\right\|_2}{\left\|u\left(t^{(k)},x\right)\right\|_2},
\end{equation}
for $k=1$. From these results, we can infer that the choice of the time step is directly related to the accuracy of the model estimation, and the number of mesh samples influences the number of training epochs. The choices of $h_t=10^{-3}$ and $n_s=100$ have provided a good balance between accuracy and computational performance.

\begin{table}[h!]
  \centering
  \caption{Problem 1. Model $1-30\times 3-1$ training tests of the first time step.}
  \vspace{12pt}
  \begin{tabular}{c|cccc}\toprule
    $h_t$\textbackslash $ns$ & $10$ & $100$ & $1000$ \\\midrule
    $10^{-1}$ & $4\e{+4}~/~2\e{-1}$ & $2\e{+4}~/~2\e{-1}$ & $2\e{+4}~/~2\e{-1}$ \\
    $10^{-2}$ & $2\e{+4}~/~3\e{-3}$ & $2\e{+4}~/~3\e{-3}$ & $2\e{+4}~/~3\e{-3}$ \\
    $10^{-3}$ & $2\e{+4}~/~3\e{-5}$ & $2\e{+4}~/~8\e{-5}$ & $3\e{+4}~/~8\e{-6}$ \\\bottomrule
  \end{tabular}
  \label{tab:prob1_fs}
\end{table}

\begin{figure}[H]
  \centering
  \includegraphics[width=0.49\textwidth]{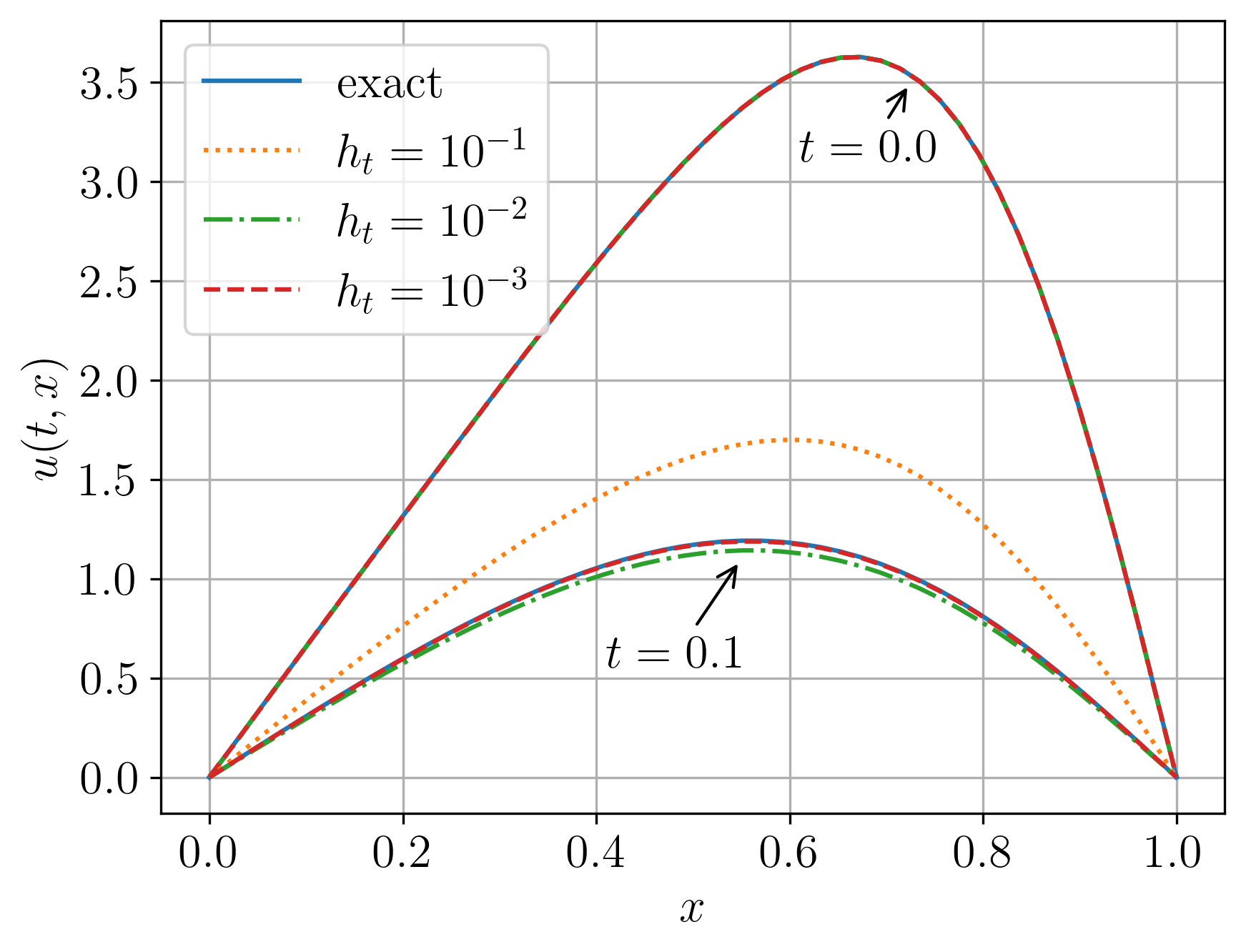}~
  \includegraphics[width=0.49\textwidth]{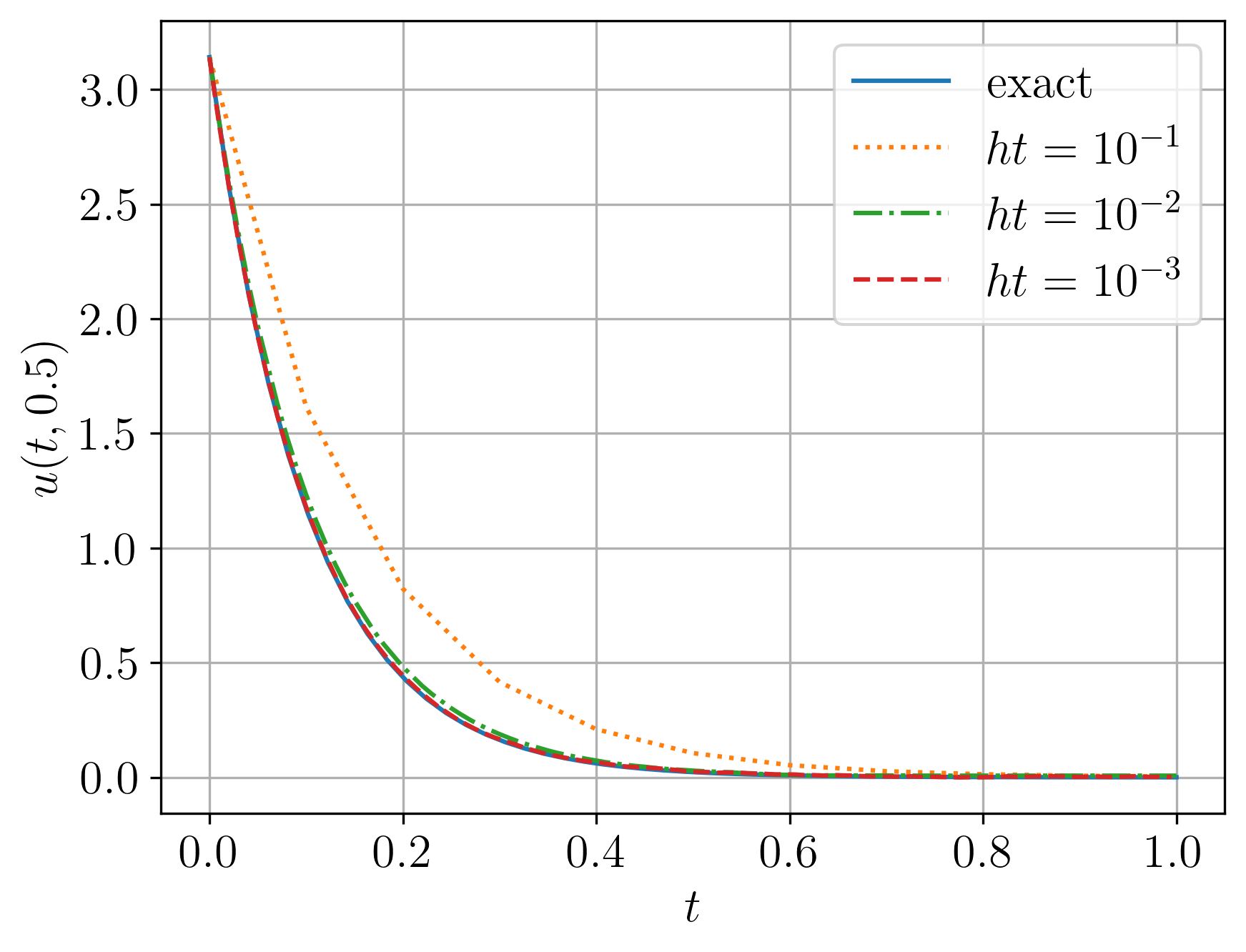}
  \caption{Numerical ($n_s=100$) \textit{versus} exact solutions for Problem 1. Left: solutions at $x=0.0$ and $t=0.1$. Right: $u(t,0.5)$ solution values.}
  \label{fig:prob1_numXexact}
\end{figure}

Figure~\ref{fig:prob1_numXexact} shows plots of the numerical ($n_s=100$) \textit{versus} the exact (solid lines) solutions for Problem 1. On the left, the solutions are compared at $t=0.0$ and $t=0.1$ for time steps $h_t= 10^{-1}$ (dotted line), $10^{-2}$ (dash-dot line), and $10^{-3}$ (dashed line). Following the same identifications, on the right, we have the comparisons for the solutions at $x=0.5$ and $0 \leq t \leq 1$.

\subsection{Problem 2}
The second test problem has the initial condition
\begin{equation}
  u_0(x) = -\sin(\pi x),
\end{equation}
and, in the work of \cite{Basdevant1986a}, the following analytical solution is given
\begin{equation}
  u(x) = \frac{-\int_{-\infty}^{\infty}\sin(\pi(x-\eta))f(x-\eta)e^{-\frac{\eta^2}{4\nu t}}\,d\eta}{\int_{-\infty}^{\infty} f(x-\eta)e^{-\frac{\eta^2}{4\nu t}}\,d\eta},
\end{equation}
where $f(y) = e^{-\cos(\pi y)/(2\pi \nu)}$. In the following, $\nu = 0.01/\pi$ is assumed.

\begin{figure}[h!]
  \centering
  \includegraphics[width=0.49\textwidth]{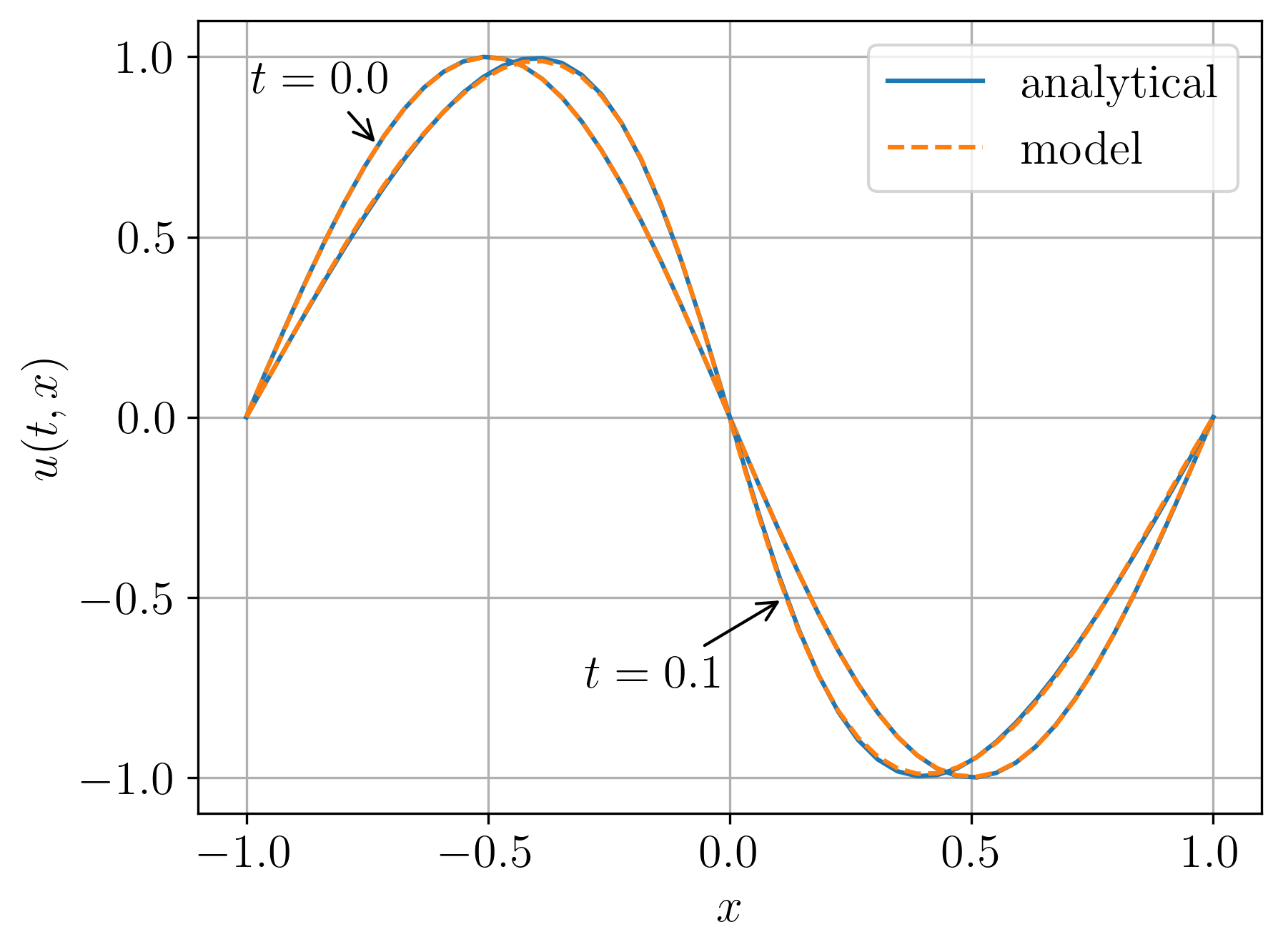}~
  \includegraphics[width=0.49\textwidth]{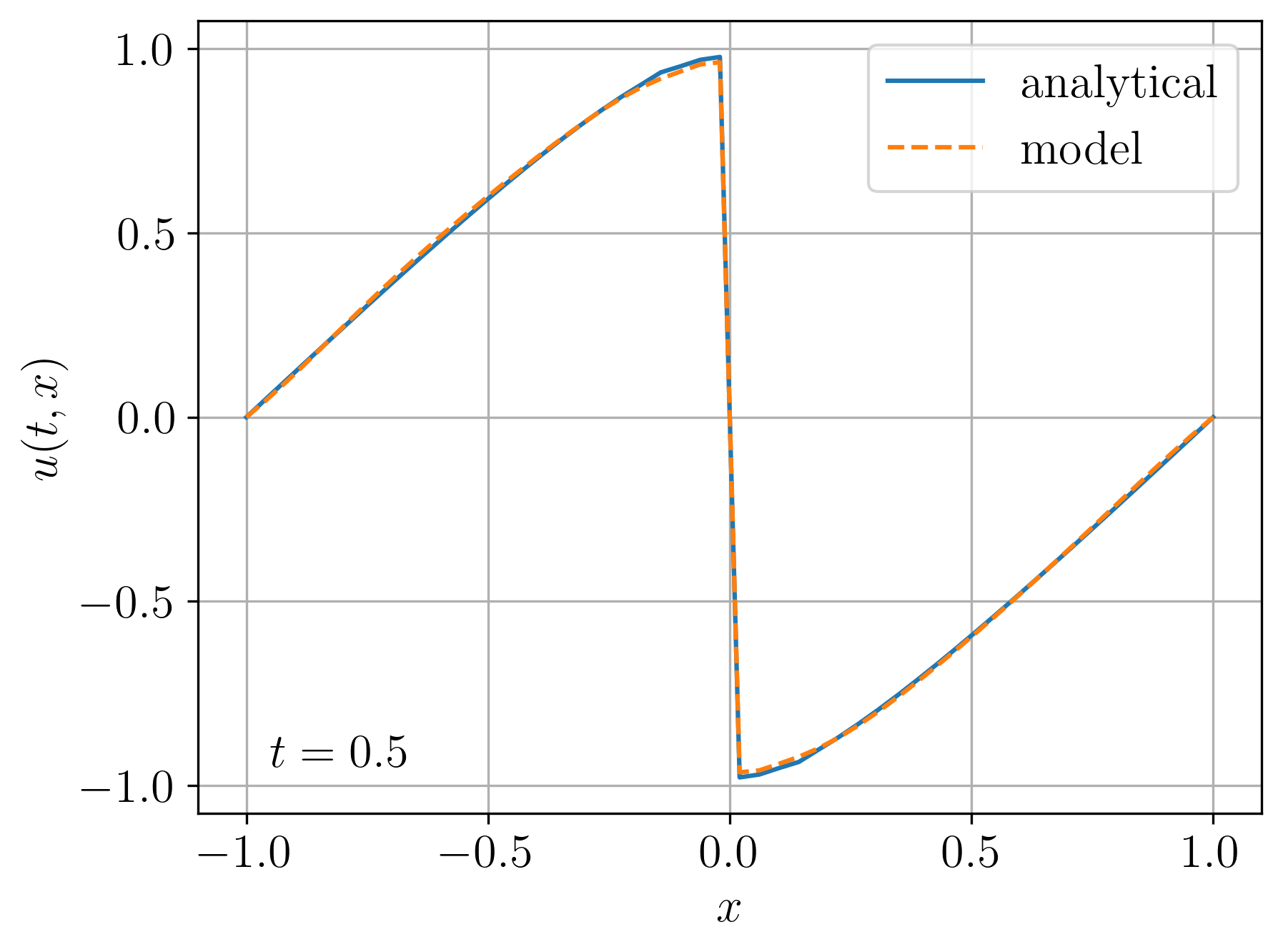}\\
  \includegraphics[width=0.49\textwidth]{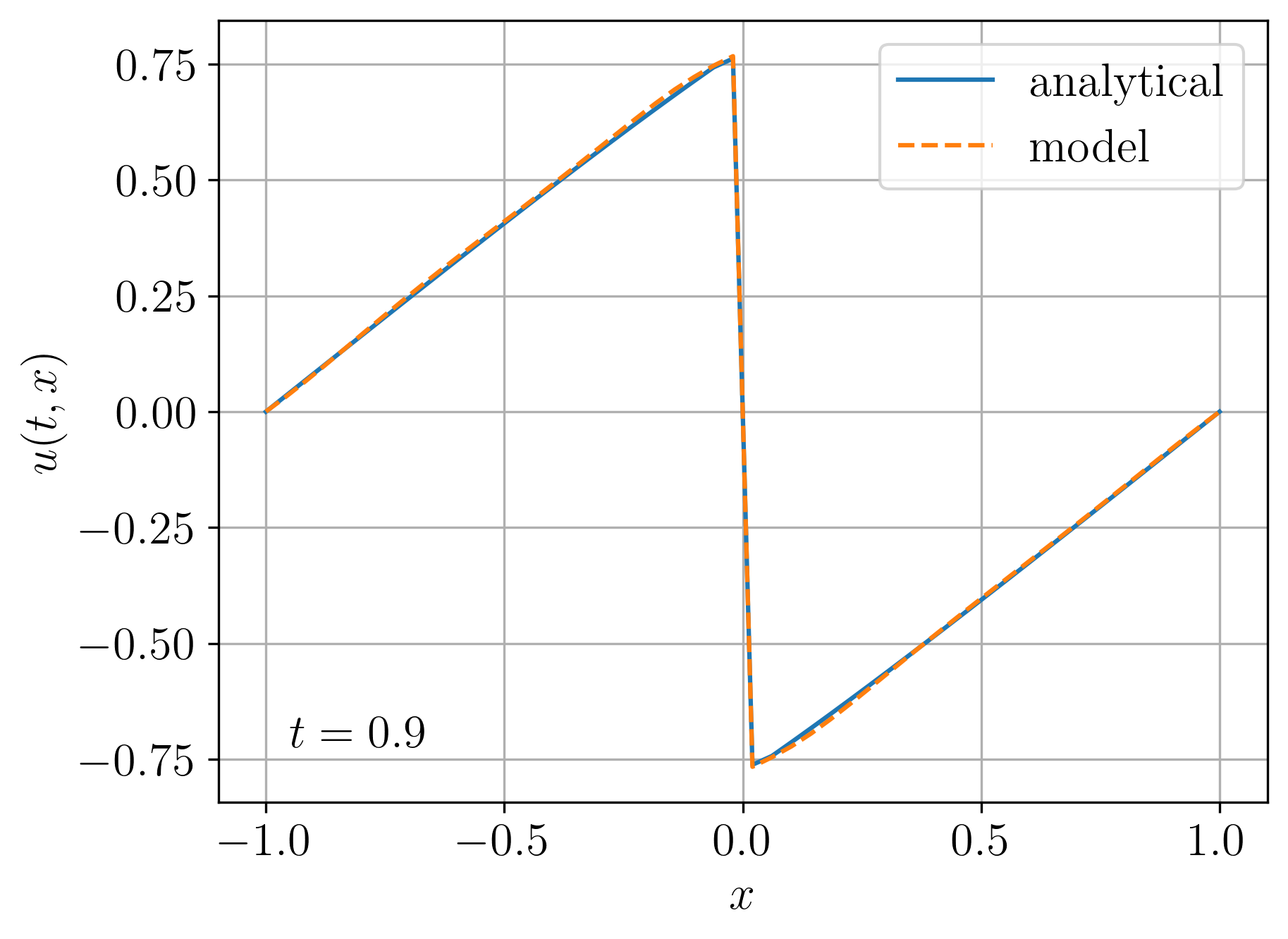}~
  \includegraphics[width=0.49\textwidth]{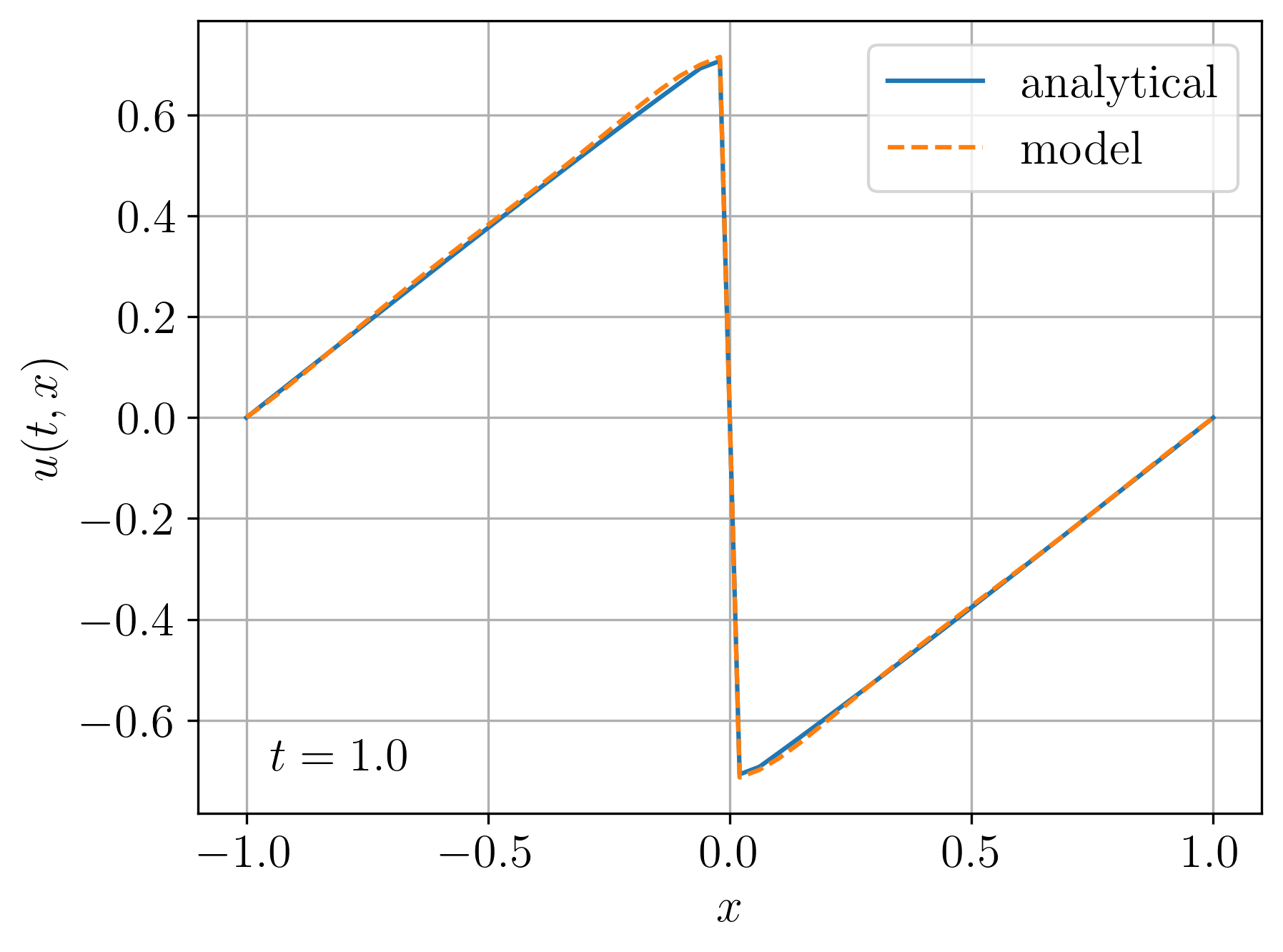}
  \caption{Problem 2. Numerical \textit{versus} exact solutions.}
  \label{fig:prob2_numXexact}
\end{figure}

By performing similar numerical tests as those discussed above, we have concluded that $1 - 30\times 3 - 1$ MLPs, with the parameters $h_t = 10^{-3}$ and $n_s=100$ were enough to produce good results. Figure~\ref{fig:prob2_numXexact} shows comparisons between model estimates (dashed lines) and analytical (solid lines) solutions at several times $t = 0.0$, $0.1$, $0.5$, $0.9$ and $1.0$. We observe that the transfer learning was efficient even in this more complex time-behavior test case.

The Problem 2 has also been studied by \cite{Raissi2019a}. By applying a PINN with input $(t, x)$ and output $u(t, x)$, they reported similar accurate results. However, their application demanded an $2 - 20\times 9 - 1$ network architecture, far bigger than the $1 - 30\times 3 - 1$ structure demanded by our proposed PINN with implicit Euler transfer learning.

\section{CONCLUSIONS}
In this work, we have proposed the application of PINNs with an implicit Euler transfer learning approach to solve the viscous Burgers equation with homogeneous Dirichlet boundary conditions. It consists of using Multilayer Perceptrons (MLPs) to estimate the solution at discrete time steps. As an iterative procedure, a first neural network model is trained to learn the initial condition. Then, the knowledge is transferred to the next model, which learns the solution at the next time step by minimizing the residual of the implicit Euler scheme. The result is a sequence of neural network models that estimate the problem solutions at discrete time steps.

In comparison to the usual PINN models, the proposed approach has the advantage of requiring smaller neural network architectures with similar accurate results and potentially decreasing computational costs. For problems with more complex dynamics, further work can include the auto-adaption of the neural network architecture during time steps. Other aspects of further developments can be its extension to other time-step schemes, like higher order Runge-Kutta schemes or multi-step schemes.

\subsection*{\textit{Acknowledgements}}
This study was financed in part by the Coordenação de Aperfeiçoamento de Pessoal de Nível Superior – Brasil (CAPES) – Finance Code 001.

% ------------------------------------------------------------------------

\end{document}